\documentclass[letterpaper, 10 pt, conference]{ieeeconf}
\usepackage[T1]{fontenc}
\usepackage{algorithm}
\usepackage{xcolor}
\usepackage{algpseudocode}
\usepackage{ upgreek }
\usepackage{multicol}
\usepackage{ dsfont,setspace }
\usepackage{amsmath,graphicx}
\usepackage{amssymb}
\usepackage{amsmath}

\usepackage{lineno}
\usepackage{epstopdf}
\usepackage{authblk}
\usepackage{dblfloatfix}
\usepackage{tabularx, makecell}
\newcolumntype{L}{>{\raggedright\arraybackslash}X}
\usepackage[
  separate-uncertainty = true,
  multi-part-units = repeat
]{siunitx}

\usepackage{subcaption}
\usepackage[colorlinks]{hyperref}
\hypersetup{
     colorlinks,
     linkcolor=black,
     citecolor=black,
     filecolor=black,
     urlcolor=blue,
 }
\usepackage{cleveref}
\usepackage{textcomp}
\usepackage{gensymb}
\usepackage{multirow}
\usepackage{svg}
\usepackage{graphicx}

\usepackage{pifont}

\IEEEoverridecommandlockouts                            
\begin{document}

\title{\LARGE \bf
DecAP : Decaying Action Priors for Accelerated Imitation Learning of Torque-Based Legged Locomotion Policies
}

\author{
Shivam Sood$^{1}$, Ge Sun$^{2}$, Peizhuo Li$^{2}$, Guillaume Sartoretti$^{2}$

\thanks{$^{1}$S. Sood is with the Department of Mechanical Engineering, Indian Institute of Technology, Kharagpur.
\href{shivamso@andrew.cmu.edu}{shivamsood2000@gmail.com}}
\thanks{$^{2}$G. Sun, P. Li, and G. Sartoretti are with the Department of Mechanical Engineering, National University of Singapore, 117575 Singapore. 
(\href{sunge@u.nus.edu}{sunge@u.nus.edu}, \href{E0376963@u.nus.edu}{E0376963@u.nus.edu} , \href{mpegas@nus.edu.sg}{mpegas@nus.edu.sg})  }
}

\maketitle
\thispagestyle{empty}
\pagestyle{empty}

\begin{abstract}
Optimal Control for legged robots has gone through a paradigm shift from position-based to torque-based control, owing to the latter's compliant and robust nature.
In parallel to this shift, the community has also turned to Deep Reinforcement Learning (DRL) as a promising approach to directly learn locomotion policies for complex real-life tasks.
However, most end-to-end DRL approaches still operate in position space, mainly because learning in torque space is often sample-inefficient and does not consistently converge to natural gaits.
To address these challenges, we propose a two-stage framework. In the first stage, we generate our own imitation data by training a position-based policy, eliminating the need for expert knowledge to design optimal controllers. The second stage incorporates decaying action priors, a novel method to enhance the exploration of torque-based policies aided by imitation rewards.
We show that our approach consistently outperforms imitation learning alone and is robust to scaling these rewards from 0.1x to 10x.
We further validate the benefits of torque control by comparing the robustness of a position-based policy to a position-assisted torque-based policy on a quadruped (Unitree Go1) without any domain randomization in the form of external disturbances during training.\footnotemark[3] 

\end{abstract}

\section{INTRODUCTION}

Legged robots excel in navigating rough terrain and cluttered areas by selecting discrete contact points. However, this agility comes at the cost of complex control challenges due to their underactuation and nonlinear dynamics. 
In addressing these challenges, Optimal Control techniques such as Model Predictive Control (MPC) have proven effective in stabilizing ground reaction forces (GRFs)~\cite{Carlo2018DynamicLI} and in achieving Whole-Body torque control~\cite{dantec2022whole}. 
Alternatively, recent advancements in Deep Reinforcement Learning (DRL) show great promise in solving these control issues. Both model-based~\cite{Sun2021OnlineLO} and model-free~\cite{rudin2022learning} DRL techniques have been applied for system dynamics improvement and end-to-end joint-level control.
At the actuator level, control mechanisms usually fall into one of two categories. The first involves converting joint angles into torque values, often facilitated by a PID controller. The second entails the direct calculation of motor torques, commonly through converting optimal GRFs. This latter method is prevalent in state-of-the-art Optimal Controllers due to the compliant and robust nature of torque-based control~\cite{Carlo2018DynamicLI, Ding2020RepresentationFreeMP}. In contrast, most model-free DRL methods predominantly operate in joint-position space, which is considered more sample-efficient for learning locomotion tasks~\cite{Margolis2022WalkTW, Hwangbo2019LearningAA}.
However, the inherent compliance of torque-based control presents a unique opportunity for more resilient and safe interactions with the environment and may offer a superior action space for DRL approaches as well.

Deploying Deep Reinforcement Learning (DRL) policies on real-world hardware poses significant challenges, primarily due to the substantial gap in sim-to-real transfer. Position-based policies often grapple with issues such as inaccurate joint angle tracking and unexpected obstacles, resulting in excessive torques and a loss of robot stability~\cite{Buchli2009CompliantQL}.
DRL policies focusing on torque exhibit a more compliant and robust nature~\cite{Fuchioka2022OPTMimicIO, Shirwatkar2023ForceCF}. However, these policies face challenges due to the sample inefficiency of the torque landscape during training. Consequently, they usually take a very long time to converge, and even then, they do not consistently converge to high-quality gaits after training. Overcoming these limitations often necessitates multiple iterations of parameter tuning and additional shaping rewards, making it a resource-intensive process.
In this paper, we aim to answer the following question: \emph{``Can we leverage the inherent sample efficiency of position-space learning to accelerate the training of an end-to-end, torque-based policy while ensuring consistent convergence to high-quality gaits?''}

\begin{figure}[t!]
\centering
\includegraphics[width=1.0\linewidth]{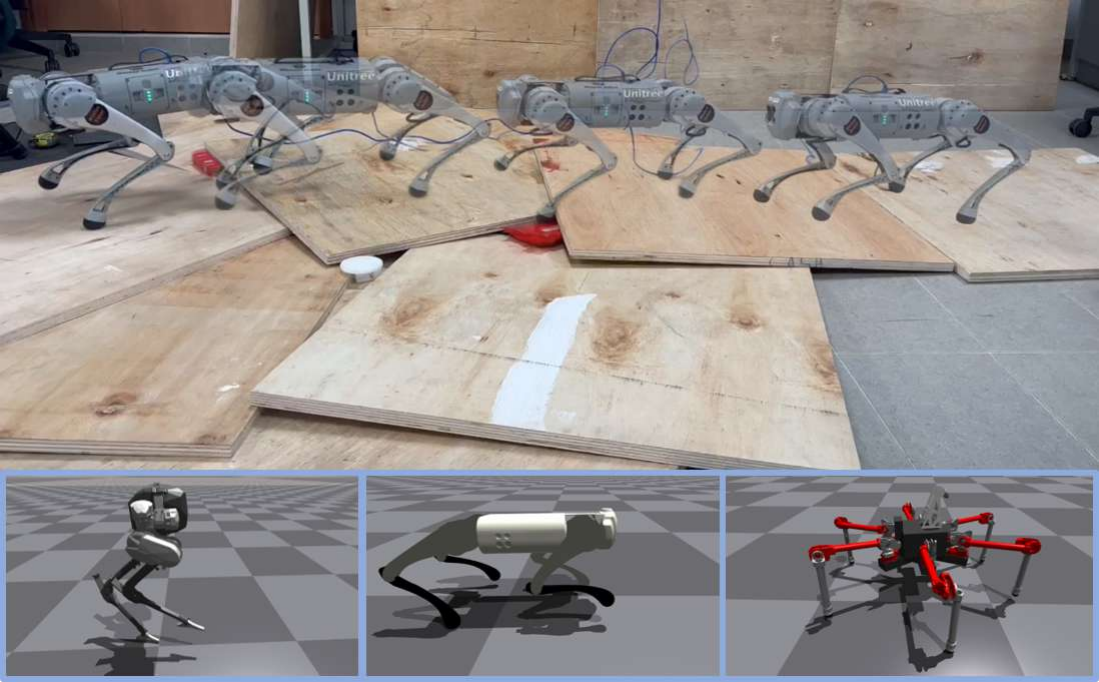}
\vspace{-1.0mm}
\caption{\textit{(Top) Our position-assisted, torque-based policy, can successfully help the robot navigate real-life uneven terrain, despite having been trained on flat ground without any external force disturbances.
(Bottom) Our torque-based velocity tracking policies for Agility-Cassie (left), Unitree-Go1 (middle), and Hebi-Daisy (right), achieve high-quality gaits within just 25 minutes of wall-clock time.}$^\textrm{4}$}
\label{fig:front}
\vspace{-6.9mm}
\end{figure}

\footnotetext[3]{Codebase: \href{https://github.com/marmotlab/decaying_action_priors}{https://github.com/marmotlab/decaying\_action\_priors}}
\footnotetext[4]{Accompanying video: \href{https://youtu.be/O1lcry7sHNQ}{https://youtu.be/O1lcry7sHNQ} }
We propose a two-stage approach, where first, we acquire our own position imitation data by training a position-based policy.
This approach is different from teacher-student frameworks like~\cite{Wu2023LearningRA, Kim2023NotOR} as our student learns in a different action space.
Nonetheless, our experiments show that solely relying on imitation learning using position data does not enhance sample efficiency for torque-space learning. 
To tackle this challenge, in the second stage, we propose Decaying Action Priors (DecAP), which guides the initial exploration of the joint-torque space by introducing torque biases calculated through a PID controller on the imitation angles.
While not optimal, these provide a beneficial initial bias, helping guide/constrain the policy during early exploration.
These biases are added to the actions sampled by the torque-based policy and vanish over time; by the end of training, the robot is able to sustain its own locomotion, without the need for such ``clutches''.

This overall approach can be explained through a ludic example, in which a newly created legged robot named Forrest is eager to learn how to run.
Having no experience in coordinating his leg joints, Forrest's initial exploration may involve erratic leg movements and occasional injuries. 
Instead, let us equip Forrest with an exoskeleton pre-programmed to mimic a running motion tailored to his legs.
At the start of his training, the exoskeleton does most of the work, allowing Forrest to focus on stabilizing himself amid potential environmental disturbances.
    As time passes, the exoskeleton weakens; however, by then, Forrest may have learned to compensate for the waning support, encouraged by our repeated shouts of ``Run Forrest, run!''.
Eventually, the hope is for Forrest to break free from the exoskeleton and be able to run independently, thanks to this experience.


Our results show that the DecAP approach significantly accelerates learning in the torque space, enabling different types of robots to complete the velocity tracking task within 25 minutes of wall-clock time, starting from scratch. DecAP consistently outperforms imitation-based approaches, maintaining notably lower root mean square errors across a wide range of imitation reward scales. 
To assess the advantages of torque as an action space we compare a position-based policy to a position-assisted torque-based policy.
Both position-based and torque policies are trained on flat terrain, without external disturbance forces during training. In real-world testing, the position-based policy failed under even minor disturbances, while the torque-based policy proved to be quite robust in out-of-distribution environments.

\section{RELATED WORK}
\label{chap:related}
There has been a recent surge in the use of RL-based control for legged robots due to its robustness to external disturbances and the ability to leverage full-body dynamics.
Successful implementations span both simulations \cite{article} and demonstrate its efficacy on hardware \cite{Hwangbo2019LearningAA, Kohl2004PolicyGR, Haarnoja2018LearningTW, 10167528}.
 Facilitated by highly parallelized DRL techniques ~\cite{rudin2022learning, Margolis2022WalkTW, Wu2023LearningRA}, the wall-clock time of training policies has significantly reduced. 
These DRL strategies are broadly categorized into two groups: model-based~\cite{article} and model-free~\cite{6315022}, the latter being the focus of this paper. Model-free approaches seek to develop end-to-end robot control policies, eliminating the need for expert knowledge required to design additional controllers to operate the policies on hardware.
In the realm of model-free RL for legged robots, Imitation Learning provides an alternative approach in which the robot learns directly from demonstrations, instead of solely relying on a reward function. 
Some studies~\cite{Fuchioka2022OPTMimicIO,Shirwatkar2023ForceCF} integrate the torque outputs of optimal control strategies into their imitation reward structures to enhance policy performance.
However, acquiring high-quality imitation data remains resource-intensive. This involves utilizing optimal control techniques, either on physical robots or in simulated environments ~\cite{Fuchioka2022OPTMimicIO, Shirwatkar2023ForceCF}, to collect such data. 
This data is then employed within a motion imitation RL framework to develop more robust control policies. 
In parallel, other research explores motion capture datasets of biological organisms \cite{Peng2020LearningAR, 8461237, 10246373} and maps the recorded joint angles to corresponding robot joints.

Unlike state-of-the-art optimal controllers~\cite{10160896} and the various imitation learning frameworks discussed above, models that do not employ imitation learning typically operate in position space, owing to its sample-efficient nature for exploration during the learning process.
Having easy-to-define stable references in position space (such as the standing position) around which stable locomotion gaits are easier to find is a major contributing factor. 
Previous comparative studies like~\cite{Peng2016LearningLS} have explored various action spaces, including position, velocity, torque, and muscle actuation, particularly in gait-cycle imitation tasks related to planar walking.
This study validates that position policies exhibit faster learning compared to torque policies and that learning in torque-space may not converge to smooth behaviors. 
However,~\cite{kim2023torque, Chen2022LearningTC} train an end-to-end locomotion policy in torque space, demonstrating its robustness over position control through various experiments. 
To the best of the authors' knowledge, only these two studies address end-to-end joint-torque control for legged locomotion. Specifically, ~\cite{kim2023torque} utilizes imitation data from an expert controller, while ~\cite{Chen2022LearningTC} introduces additional shaping rewards tailored to a quadruped configuration.
The robust hardware behaviors observed in these studies show that, if the sample efficiency of torque-space learning can be addressed, it would offer a superior action space for end-to-end joint control policies. 

\begin{figure*}[ht] 
    \vspace{1.7mm}
	\centering
	\includegraphics[width=0.85\linewidth]{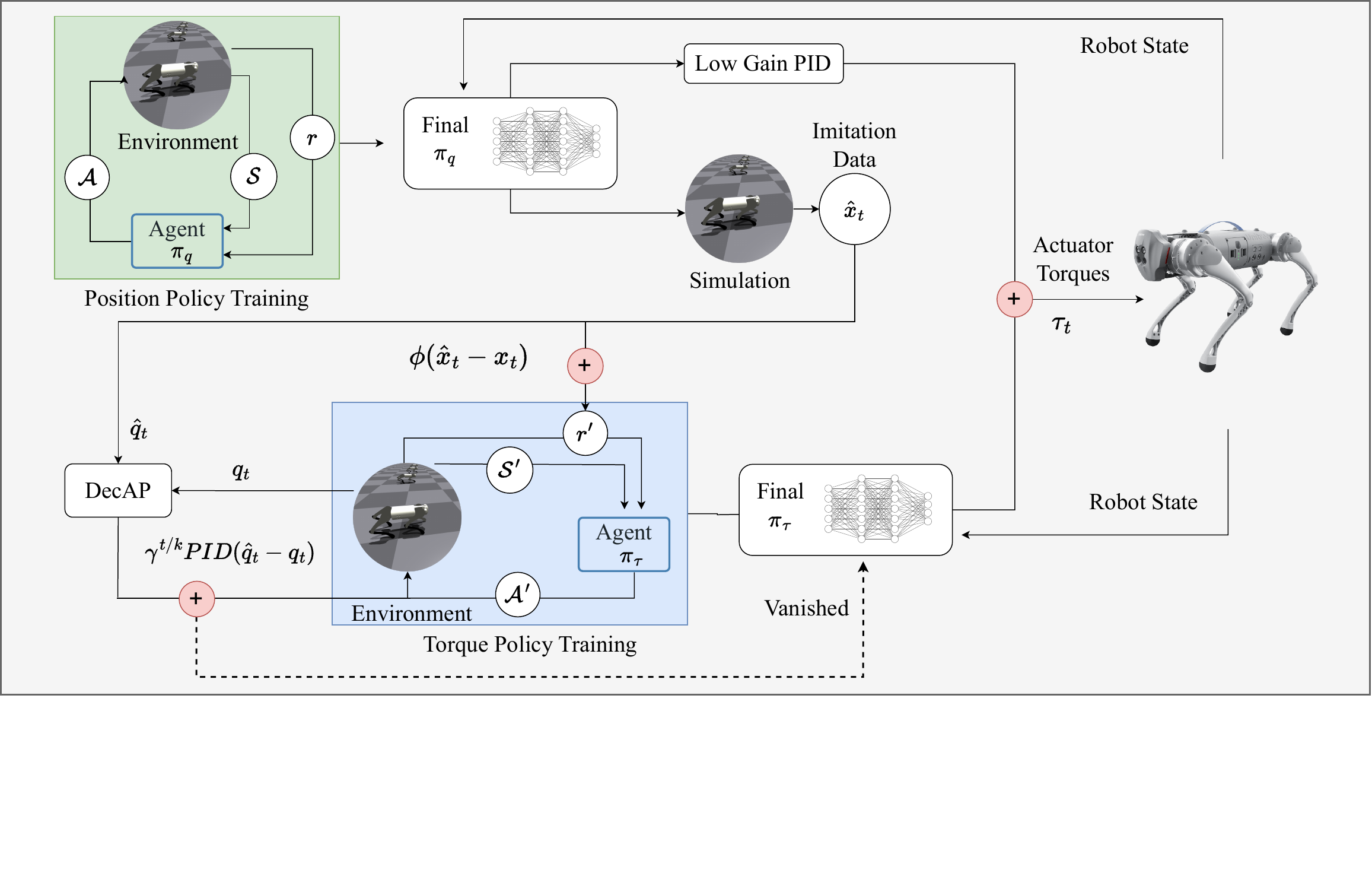}
 \vspace{-19.5mm}
	\caption{ Overview of the proposed torque learning framework: First, we train a position-based policy $\pi_q$, to acquire offline position imitation data ($\hat{x}_t$) for robot state ($x_t$), which is incorporated into the reward structure while training the torque-based policy. At the same time, we augment the sampled actions ($\mathcal{A}'$) with a torque bias ($PID(\hat{q}_t-q_t)$). This torque bias, calculated from the joint-position imitation data, guides the initial actions for faster convergence and is multiplied by a gradual time decay factor $\gamma^{t/k}$. Finally, after the torque bias becomes negligible and the torque-based policy's actions alone are sufficient to operate the robot, we deploy these torques along with a low-gain PD controller to send the final torques ($\tau_t$) to the robot actuators.}
	\label{fig:pipeline}		
    \vspace{-3,5mm}
\end{figure*}

\section{Methodology}
In this section, we formalize the base velocity-tracking control problem in the context of continuous state and action space RL. We detail our approach (Fig.~\ref{fig:pipeline}), which shows the stages of the DecAP framework: the collection of imitation data, decaying action priors for accelerated learning in the torque space, and the deployment of our position-assisted torque-based policy on hardware.

\label{chap:training_framework}
\subsection{Problem Formulation}
We frame our control problem as a Markov Decision Process (MDP) denoted as a tuple $\mathcal{M}$ = $[\mathcal{S}, \mathcal{A}, \mathcal{T}, r, \gamma]$ where $\mathcal{S}$ is the state space, $\mathcal{A}$ is the action space, $\mathcal{T}(s_{t+1}|s_t,a_t)$ are the transition probabilities that describe the dynamics of the system, $r(s_t,a_t,s_{t+1})$ describes reward received when transitioning to state $s_{t+1}$ from $s_t$ after taking action $a_t$ and $\gamma$ is the discount factor. For our legged locomotion task, the objective is to learn a policy $\pi : \mathcal{S} \to \mathcal{A}$ that maximizes the cumulative discounted sum of all future rewards, represented by the expected return $J(\pi)$:\\
\begin{equation}
\vspace{-1mm}
    J(\pi) = \mathbf{E}[\sum_t^{\infty} (\gamma^t r(s_t, \pi(s_t), s_{t+1}))]
\vspace{-4mm}
\end{equation}

\begin{table}[htbp]
\label{shaping_rewards}
	\centering
 \resizebox{\linewidth}{!}{
	\begin{tabular}{ c c c }
		\hline
		\textbf{Reward term } 		& \textbf{Expression} & $\textbf{w}$ \\
		\hline
              linear velocity	& $\phi(v^{cmd}_t - v_t)$	& $1dt$   \\
            angular velocity		& $\phi(\omega^{cmd}_t - \omega_t)$	& $1dt$   \\
		collisions			& $-n_{collision}$ & $1dt$	\\
		action rate		& $-||\dot{q}_t^*||$	& $0.01dt$\\
		orientation		& $-||\Phi||^2-||\theta||^2$	& $5dt$ \\
		angular velocity penalty		& $||\omega_{xy}||^2$	& $0.05$ \\
		linear velocity penalty		& $-v_z^2$	& $2dt$\\
  		joint torques		& $-||\tau||^2$	& $10^{-5}dt$ \\
        joint motion		& $-||\Ddot{q_t}||^2 - ||\dot{q_t}||^2$	& $2.5 \times 10^{-7}dt$ \\
        feet slip		& $||v^{foot}_{xy}||$	& $0.04dt$  \\
		\hline
	\end{tabular}}
 	\caption{\textit{${R}_{tsk}$ and weights ($\textbf{w}$). $\phi{(x)}$ represents the squared exponential $e^{-||x||^2/\sigma}$  with $\sigma=0.25$, $v^{cmd}_t$ and $\omega^{cmd}_t$ are the commanded linear and angular velocities, $v_t$ and $\omega_t$ are base linear and angular velocities, $n_{collision}$ is the number of collisions, $\dot{q}_t^*$ are desired joint angles, $\Phi$ and $\theta$ are base roll and pitch angles,  $\omega_{xy}$ is the base angular velocity in $xy$ plane, $v_z$ is the linear velocity in $z$ direction, $\tau$ is the commanded torque, $\Ddot{q}_t$ and $\dot{q}_t$ are the joint acceleration and joint velocity and $v_{xy}^{foot}$ is the velocity of foot in $xy$ plane.} }
	\label{shaping_rewards}
    \vspace{-6mm}
\end{table}

\subsection{DecAP framework}

\subsubsection{State and Action Space}
For our state, we leverage the observations from the proprioceptive sensors, including the IMU and joint encoders. The state space $\mathcal{S} = [g_{proj},v^{cmd}_{t}, \omega_{t}^{cmd}, q_t, \dot{q}_t, a_{t-1}]$ consists of the robot's base projected gravity vector  $g_{proj}$, user velocity commands  $v^{cmd}_{t}$ and $\omega_{t}^{cmd}$ (consisting of the $x$ and $y$ components of linear velocity and the angular velocity command), the motor positions $q_t$, motor velocities $\dot{q}_t$ and one time-step history of the unscaled actions $a_{t-1}$ sampled from the policy.
We use the same states for our torque and position policies, ensuring a fair comparison.
The action space for both policies is a vector with the same dimension as the number of actuators.
These actions are scaled by a constant factor (action scale), which we empirically set to be $8.0$ for training the torque policies and $0.25$ for the position policies on each of the robots in simulation.

\subsubsection{Position-based Policy Training}
In order to obtain imitation data essential for faster learning in torque space, we initially train an end-to-end joint position-based policy.
We employ straightforward high-level rewards and regularization rewards similar to the approach outlined in~\cite{rudin2022learning} and utilize the PPO-Clip algorithm~\cite{article} for training.
These rewards $R_{tsk}(s_k,a_k,s_{k+1})$ are shown in Table~\ref{shaping_rewards}.
Highly parallelized learning enables faster learning of position policies and multiple reward-tuning iterations to achieve high-quality gaits. 
While similar methodologies like \cite{Fuchioka2022OPTMimicIO, Peng2020LearningAR} can be applied within our framework, this approach eliminates the expert knowledge required to design optimal controllers in these approaches and allows us to collect robot-specific imitation data.

\subsubsection{Collection of Imitation Data}
\begin{figure}[h]
\centering

\includegraphics[width=\linewidth, height=1.5in,trim=40 130 80 130,clip]{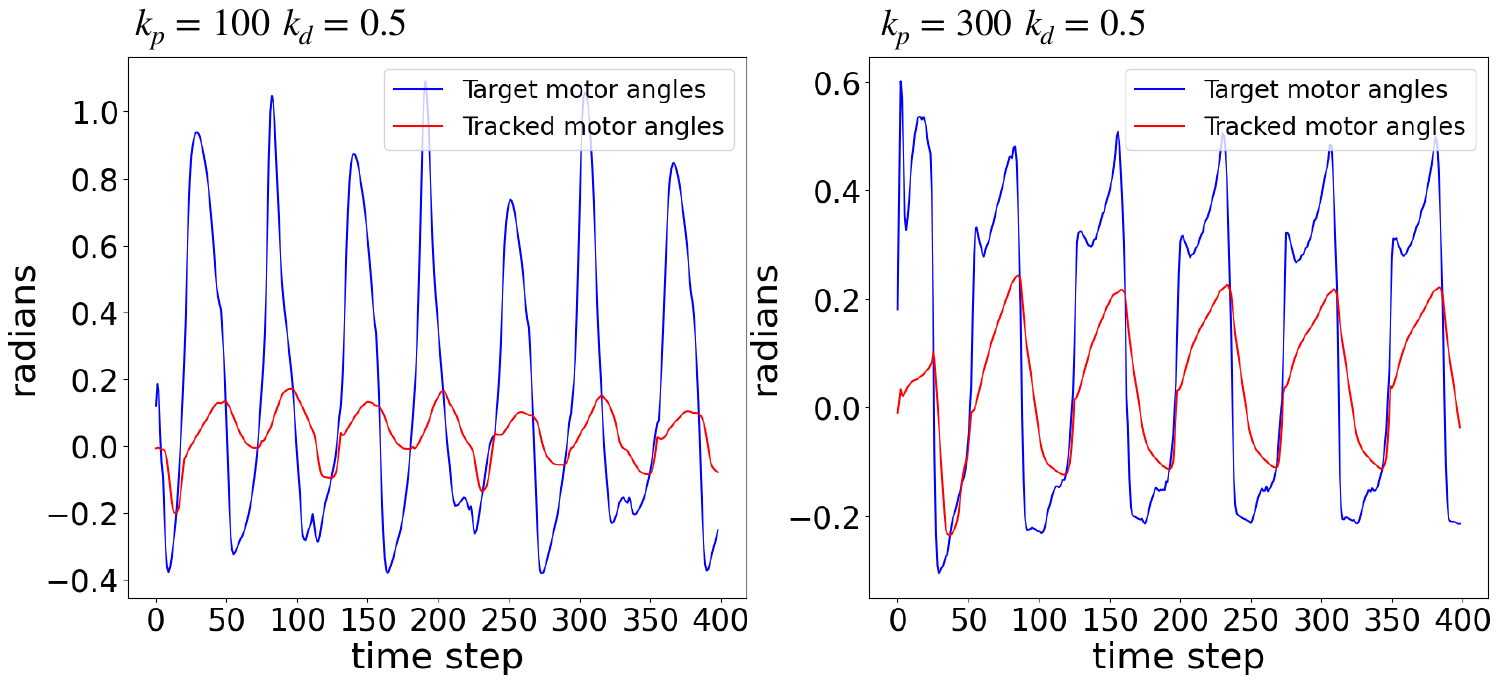}

\caption{\textit{The position-based policy generates different action outputs depending on the PID gains, while the tracked angles of the robot in simulation remain relatively stable, which makes them better suited for imitation}}
\label{pid_tracking_different_gains}
\end{figure}

Position-based policy's output actions (desired joint angles) exhibit significant sensitivity to the PID gain parameters set during its training phase, as illustrated in Fig~\ref{pid_tracking_different_gains}.
For instance, lower gains result in notable overshooting of the desired angles to ensure that the actuators closely follow the intended trajectory. 
Therefore, without extensive tuning of these gains, the policy's output will not precisely match the reference imitation joint angles we require.
To obtain more accurate imitation data, a crucial element in our framework, we leverage the RL paradigm asserting that ``simulations are doomed to succeed''~\cite{Brooks1993RealRR}. The tracked angles in the simulation are such that the reward is maximized regardless of the PID tuning values. We utilize these tracked variables as motion imitation data to train the torque-based policy.  

 Specifically, we collect the following data from the simulation: 
 $[\hat{q}_{t}, \hat{h}_{t}, \hat{r}^{e}_{t},\hat{r}^{z}_{t}, v^{cmd}_t, \omega^{cmd}_t]$, where the $\hat{q}_{t}$ are the simulation-tracked motor angles, $\hat{h}_{t}$ is the base height, $\hat{r}^{e}_{t}$  is the end-effector position in body frame, $\hat{r}^{z}_{t}$ is the foot height in world frame, $v^{cmd}_t$ and $\omega^{cmd}_t$ are the linear and angular velocity commands at each time step respectively. 

\subsubsection{Incorporating Imitation Learning}
Due to the sample-inefficient nature of exploration in torque space, the reference points provided by imitation data can help learn more natural and bio-mechanically sound motion patterns.
We use  imitation rewards($R_{im}(s_k,a_k,s_{k+1})$) based on  as~\cite{Peng2020LearningAR}:
\begin{equation}
    R_{im}(s_k,a_k,s_{k+1}) = exp \Big[\frac{-||\hat{x}_t-x_t||^2}{\sigma}\Big]
\end{equation}
where $\hat{x}_t$ is the reference state variable the system is being rewarded for imitating at time step $t$ and $x_t$ represents the corresponding state of the variable at the same time step. The standard deviation $\sigma$ controls how close to the reference point the agent has to be for a significant reward.
All our imitation rewards are framed using this squared exponential function except for the base height, which we frame as a penalty.
Along with imitation rewards, we make use of shaping rewards (Table~\ref{shaping_rewards}) as well to make sure we learn smoother motions. The total reward thus becomes $r = R_{im} + R_{tsk}$.

\begin{table}[H]
    \vspace{2.4mm}
	\centering
	\begin{tabular}{ c c c c }
            \hline
		\textbf{Reward term}  		& \textbf{Expression} & $\textbf{w}$ & \textbf{$\sigma$}\\
		\hline
		joint angles				& $\phi(\hat{q}_{t} - q_t)$	& $1.5dt$ & $0.1$\\
		end-effector position		& $\phi(\hat{r}^{e}_{t} - r^e_t )$	& $1.5dt$ & $0.1$\\
		foot height		& $\phi(\hat{r}^{z}_{t}-
        r^{z}_t)$	& $1.5dt$ &   $0.025$\\
            base height		& $-|\hat{h}_{t} -  h_t|$	& $10dt$ &   $-$\\
		\hline
	\end{tabular}
 	\caption{\textit{${R}_{im}$ and weights ($\textbf{w}$). $q_t$ are the joint angles, $r_t^e$ is the body-frame end-effector position, $r_z^t$ is the world-frame foot height, and $h_t$ is the base height.}}
	\label{imitation_rewards}
    \vspace{-5mm}
\end{table}

\subsubsection{Decaying Action Priors}
\label{pgpe}
For faster training and stable convergence to the imitation gait, we introduce an intuitive bias into the policy's sampled actions by employing a PD controller to translate the reference imitation angles into torque values:
\begin{equation}
    \beta =  K_p(\hat{q}_{t} - q_t) +  K_d(-\dot{q}_t) 
\end{equation} 
where $t$ is the current time step, $K_p$ and $K_d$ are the tuning gains controlling the bias in the exploration.
The torque values are then integrated into the torque-based policy's actions with a gradual time-decay factor:
\begin{equation}
    \tau_t = a_t \sim \pi_{\tau}(s_t) + \gamma^{t/k}( \beta )
\end{equation}
where $\tau_t$ is the torque sent to the motors, $a_t$ is the action sampled from the torque-based policy $\pi_{\tau}$, $\gamma < 1 $ and $k$ are the hyper-parameters that control the decaying speed. 
In our case, $\gamma = 0.99$ and $k=100$.
It is akin to a motor following a desired angle using PD control. While these PD torques may not perfectly align with the ideal torques for each motor, they provide an adequate initial bias. 

\begin{figure*}[h!]
\centering
\subfloat[][]{
\includegraphics[width=0.325\textwidth]{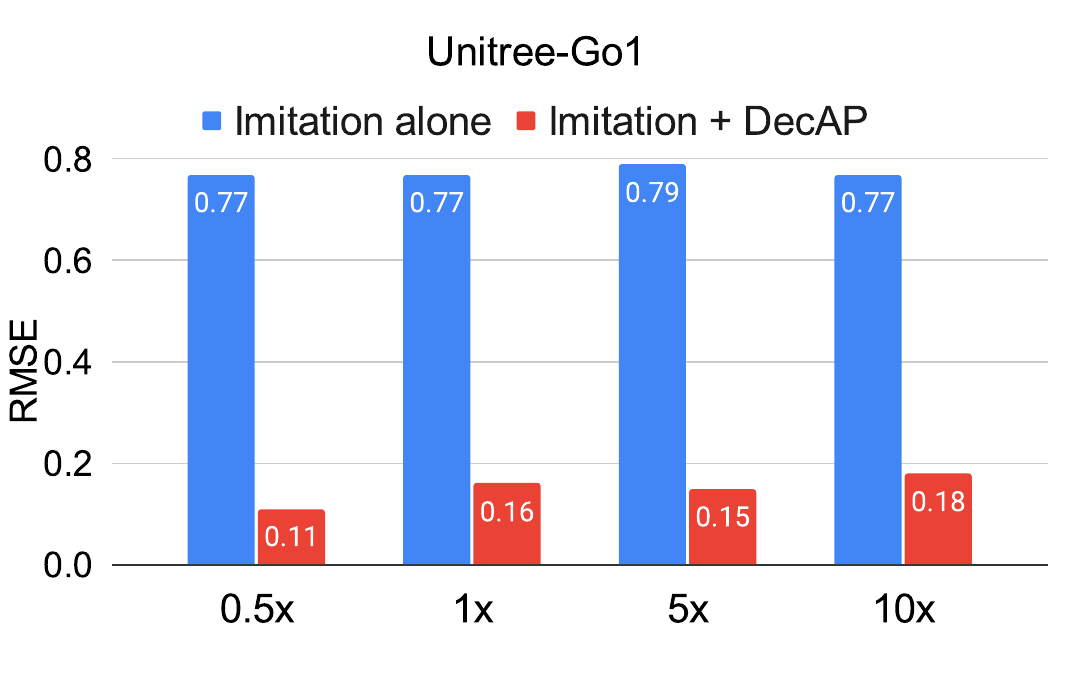}
}
\subfloat[][]{
\includegraphics[width=0.325\textwidth]{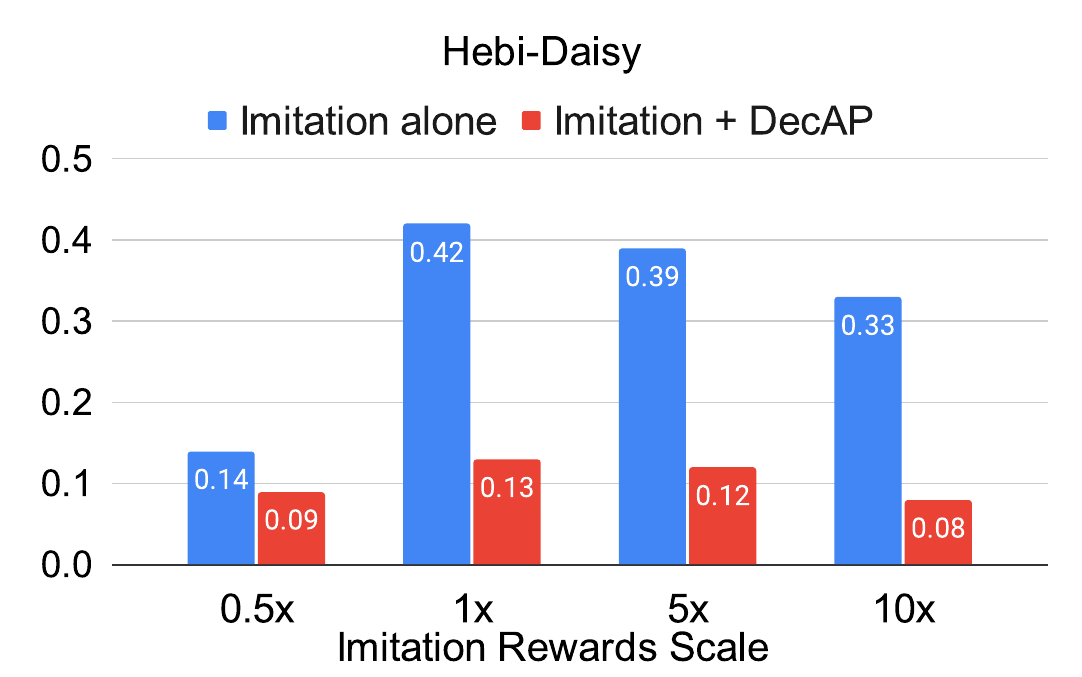}
}
\subfloat[][]{
\includegraphics[width=0.325\textwidth]{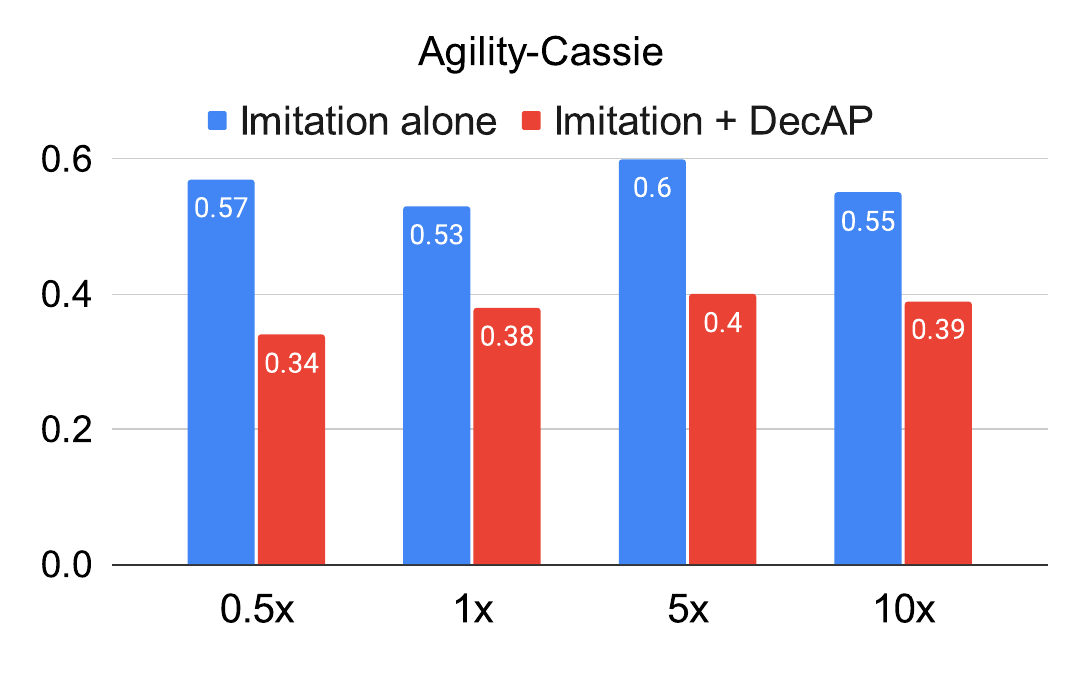}
}
\vspace{-5mm}
\caption{\textit{Comparing RMSE (in radians) between the simulated robot's tracked angles and reference imitation angles at different reward weights, DecAP + Imitation consistently outperforms imitation alone when learning in torque space. Relying on position imitation data alone yields unnatural gaits, evident from the huge deviations from reference imitation angles.}}
\vspace{-2mm}
\label{fig:decap_comparison_erros}
\end{figure*}


\subsubsection{Implementation Details}
We trained both the torque and position policies using the NVIDIA Isaac-Gym framework~\cite{Makoviychuk2021IsaacGH}.
Both training and testing of the policies is done on a workstation equipped with an i7-13700KF CPU and an RTX 4090 GPU. Training each of the position and torque policies takes about 25 minutes. Each training run had 4096 agents train for 1000 policy iterations. Both the torque and position policies run at 200Hz in simulation and on hardware. For training purposes, we verify our approach using similar PPO hyperparameters as~\cite{rudin2022learning}. Both our actor and critic policy network sizes are [512,256,128].

\section{Experiments And Results}
\label{chap:results}
This section presents both simulation and hardware experiments, where we compare factors such as learning speed, reward sensitivity analysis, gait comparison, and behavioral stability. 
For the simulation experiments, we test our approach on three different robotic platforms: Unitree-GO1, Hebi-Daisy, and Agility-Cassie. 
We then validate the trained policy on a Unitree-GO1 robot in the real world.

\subsection{Simulation Results}
\label{simulation_results}
\subsubsection{Learning Efficiency}
\begin{figure}[h!]
	\centering
\includegraphics[width=0.47\textwidth]{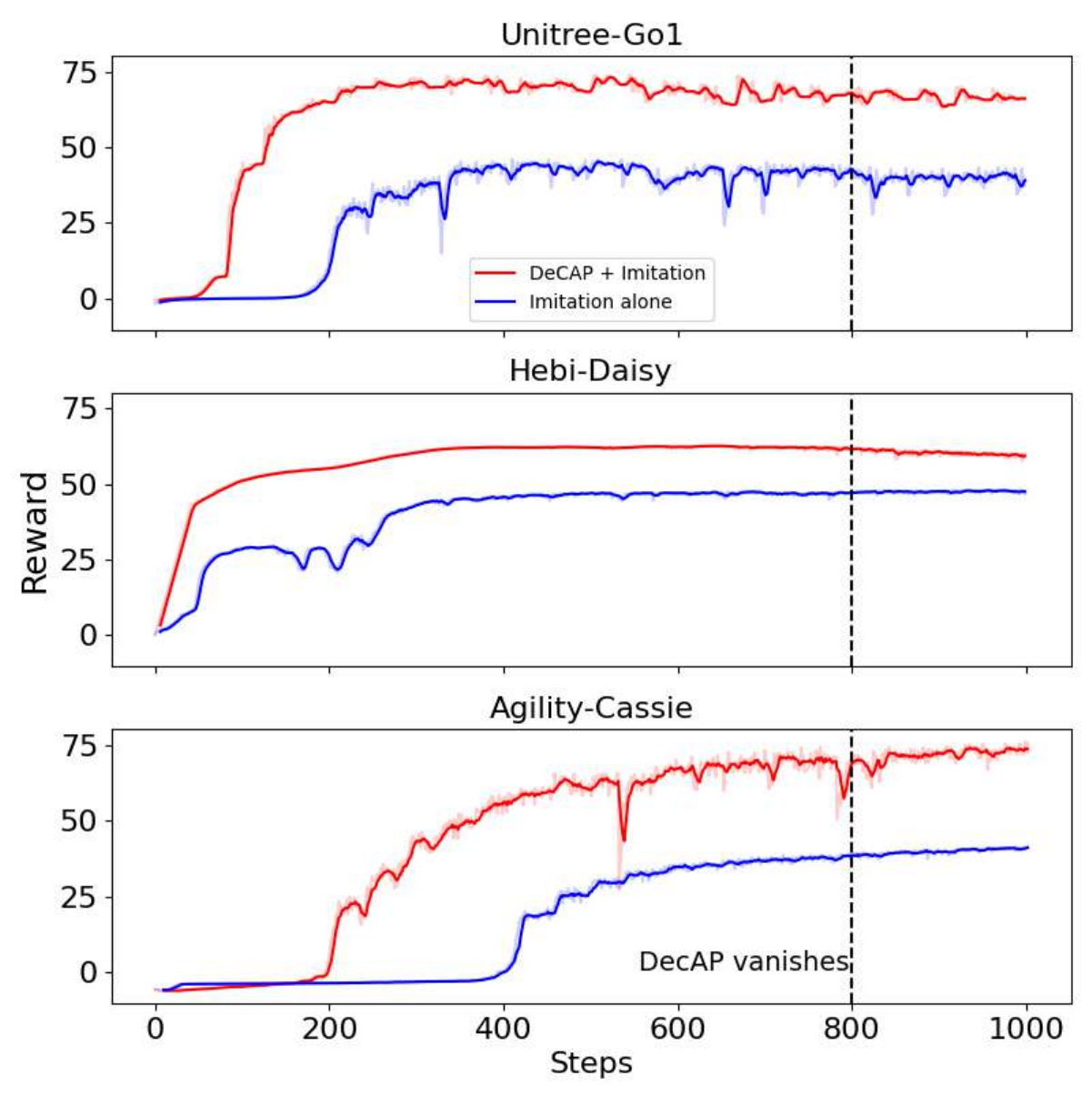}
	\caption{\textit{Comparing the reward progression over time for various legged robots, we observe that the vanishing of DecAP indicates the action priors have become insignificant. This suggests that the actions sampled directly from the policy are now adequate for controlling the robot.}}
	\label{fig:learning_time}		
\end{figure}

To ensure a fair comparison, we maintain identical reward structures and weights for both imitation learning and DecAP. Fig. \ref{fig:learning_time} shows the reward over 1000 timesteps, which corresponds to approximately 25 minutes of wall-clock time, using the specified nominal reward weights outlined in Table \ref{shaping_rewards} and \ref{imitation_rewards}. 
By timestep 800, DecAP's contribution becomes negligible, and the torques sampled from the policy suffice for robot control in simulation. 

\subsubsection{Reward Sensitivity Analysis}
\label{reward_sensitivity} 
We tested our approach across a broad range of imitation reward weights, specifically the joint angles, end-effector positions, and foot height imitation reward weights. 
The nominal value for each of these weights was set at 1.5. These weights were then multiplied by scales:$[0.5, 1, 5, 10]$ for both the imitation-only and the imitation + DecAP approaches. 
For comparison, we calculated the Root Mean Square Error (RMSE) between the reference imitation angles and the tracked imitation angles over 1000 iterations. 
As illustrated in Fig. \ref{fig:decap_comparison_erros}, DecAP significantly outperforms the imitation-only approach, while also simplifying the task of reward tuning. 
These results are best demonstrated in the attached video.
\begin{figure}[h]
	\centering
\includegraphics[width=0.48\textwidth]{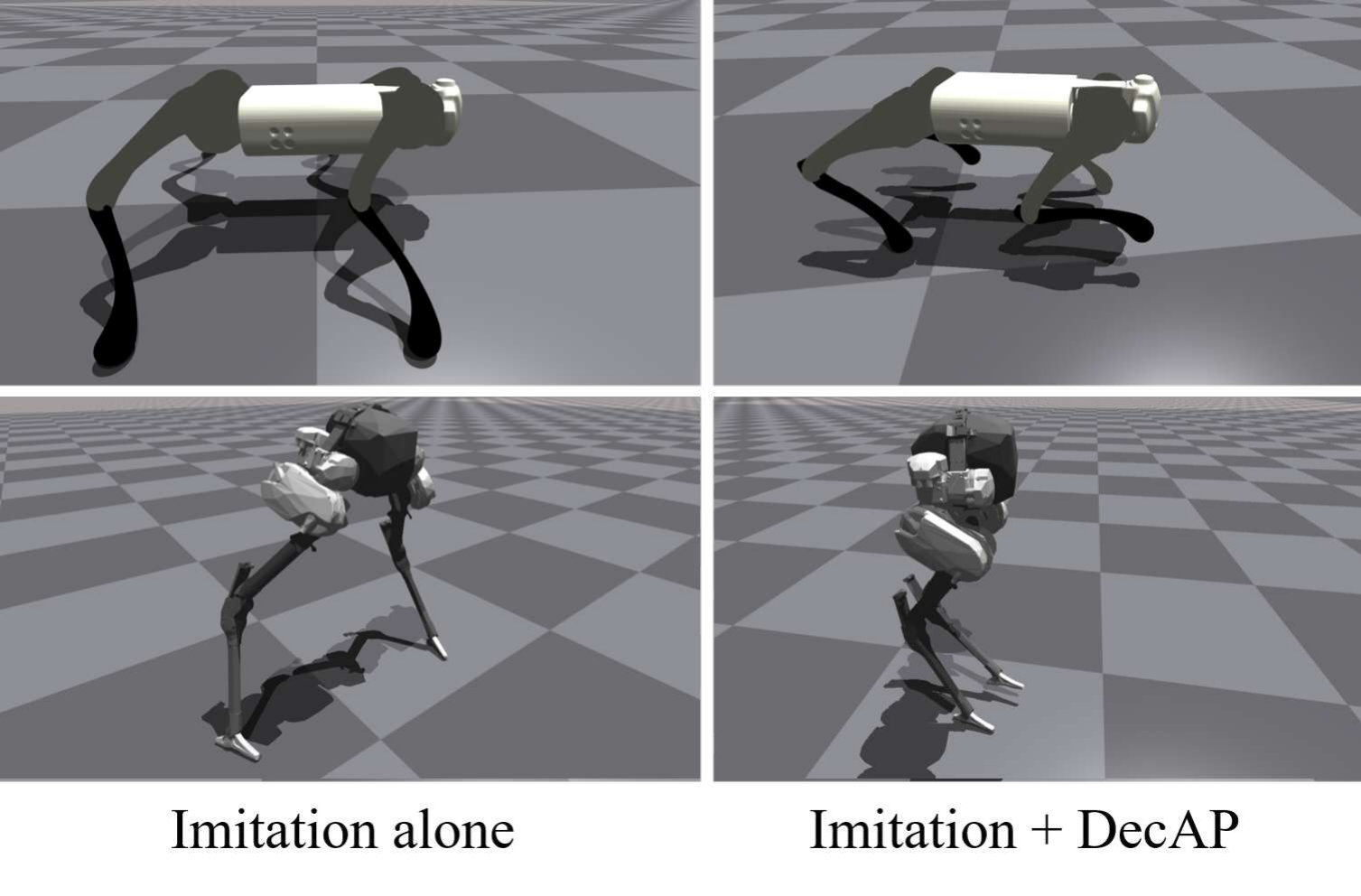}
        \vspace{-6.0mm}
	\caption{\textit{Torque-based velocity tracking policies trained using imitation alone (left) and using Imitation + DecAP (right). The DecAP framework quickly converges to a high-quality gait, while imitation alone generally converges to awkward gaits due to sample inefficiency of exploration torque space.}}
	\label{fig:stable_reference}		
	\vspace{-4.0mm} 
\end{figure}

\begin{figure*}[h!]
\centering
\subfloat[][]{
\includegraphics[width=0.85\textwidth, trim=90 100 60 120,clip]{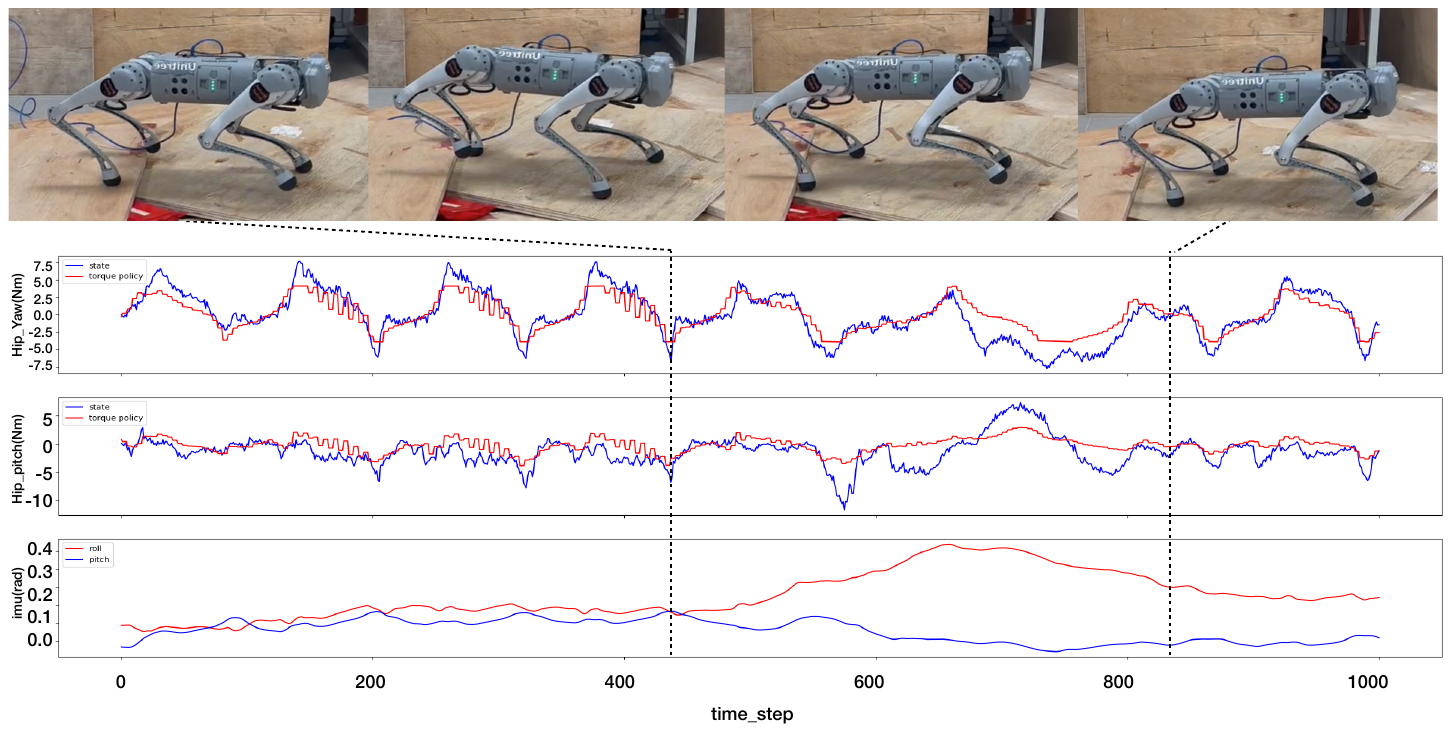}
}
 \vspace{-10mm} 

\caption{\textit{Snapshots of our hardware experiments featuring the Unitree Go1 quadruped. We deploy our positon-assisted torque-based policy trained with DecAP, with a low-gain PID on our position-based policy. The robot is tasked with traversing a non-flat, out-of-distribution terrain. The dotted line indicates the time-steps of disturbance (an uneven step). The bottom section includes the torque-based policy output, actuator torque feedback for the rear right leg, and IMU data collected during the experiment.
Note how our torque-based policy maintains a smooth output during an external disturbance and helps mitigate the position-policy PID's abrupt torque fluctuations and recover the robot's orientation.}}
 \label{fig:hardware}
 \vspace{-6.7mm}
\end{figure*}

\subsubsection{Gait comparison with and without DecAP}
A major reason for the sample inefficient nature of torque-space learning is the lack of stable reference points around which natural gaits occur (like the standing position for a quadruped). Torque-based policies typically begin with zero torques and are more likely to converge to unnatural gaits, as illustrated in Fig.~\ref{fig:stable_reference}, where the torque-based policy learns an awkward standing position on both robots. 
Decaying Action Priors address this issue by providing torques in the required direction to initiate from this position and subsequently improve imitation of reference angles. This effect is also evident in Fig.~\ref{fig:learning_time} where using only imitation learning results in a prolonged period of learning a stable reference and lower reward values. In contrast, DecAP quickly improves and stabilizes the robot.

\subsection{Hardware Experiments}
We deploy our torque-based policy on hardware using a sim-to-real approach inspired by Model Predictive Control (MPC) based approaches for legged robot control~\cite{Carlo2018DynamicLI}. 
This approach implements a low-gain Proportional-Integral-Derivative (PID) controller (on our learned position policy) in conjunction with our torque-based policy.
To demonstrate the robustness of our controller, we test both our position-based and torque-based policies on out-of-distribution, non-flat terrains.
For safety reasons, the scaled outputs are clipped when running the torque-based policy on hardware.



The robot's task is to traverse a non-flat terrain consisting of randomly placed wooden boards (Fig~\ref{fig:hardware}).
The position-based and torque-based policies, trained exclusively on flat terrain without external disturbances as domain randomization during training, demonstrate distinct behaviors. 
The torque-based policy adeptly handles disturbances, while the position-based policy struggles with instability resulting in the robot falling over, as evident in the supplementary video. 
During the analysis, we gather real-time data, including the torque-based policy's actions, actuator torque estimates, and IMU data (pitch and roll). 
Regular trotting (time steps within $[0, 450]$) shows periodic variations in both estimated torques and policy outputs, with policy outputs contributing to about 70\% of the total actuator torque. 
When facing disturbances (time steps within $[450, 830]$), actuator torques exhibit rapid changes, characterized by peaks and valleys differing from typical trotting values.
However, the torque-based policy maintains stable outputs. This suggests that the low-gain PID controller running alongside the torque-based policy generates drastic torques as it aims to achieve desired joint angles suitable for flat terrain, disregarding the encountered disturbance. This leads to abrupt torque fluctuations in some actuators, increasing system noise and instability, causing the position-based policy (if running alone) to fail. But when combined, the torque-based policy's smooth output during this phase mitigates the PID controller's abrupt changes on the total torque output, enhancing robot stability and successfully recovering from the roll deviation after timestep $830$. 
While existing RL policies struggle to perform in the real world without domain randomization (in the form of external force disturbances during training), the experiments demonstrate that our torque-based policy navigates the out-of-distribution non-flat terrain without domain randomization. We believe that this highlights the superior nature of learning a legged policy in torque space since it offers more resilient and safer interactions with the environment.

\section{Conclusion And Future Work}
\label{chap:conclusion}
We propose a framework enabling fast, end-to-end training of torque-based policies for legged robots. 
Our algorithm, DecAP, harnesses the sample efficiency of position-based learning to significantly accelerate learning in the torque space. 
We achieve this by incorporating imitation rewards derived from position-based policies and guiding action exploration in the torque space. This also makes the rewards robust to scaling compared to traditional imitation learning approaches.
With our method, torque-based policies converge to a high-quality gait exhibiting robustness in the face of external disturbances.
Our model-free approach generalizes well across various legged robots and is validated through hardware experiments with a quadruped.

Our current approach relies on offline data from position-based policy simulations, limiting it to the available imitation data.
We intend to parallelize the trained position-based policy simulations with torque-based policy training in the future to collect imitation data online. This enables us to have imitation data for all the possible commands a position-based policy has learned.
Also, instead of decaying all the action priors at the start, we aim to adopt a performance-based approach for action priors, only assisting robots as they face challenges and enabling curriculum learning for various terrains, leveraging expert position policies. 
This would enable faster torque learning on diverse terrains by incorporating exteroceptive data like the terrain map for real-world locomotion. 
\section{Acknowledgements}
 This work was supported by the Singapore Ministry of Education Academic Research Fund Tier 1, as well as by the National Research Foundation, Singapore (NRF), Maritime and Port Authority of Singapore (MPA) and Singapore Maritime Institute (SMI) under its Maritime Transformation Programme (Project No. SMI-2022-MTP-01).
\bibliographystyle{IEEEtran}
\bibliography{references}

\end{document}